\documentclass[11pt]{article}

\usepackage[final]{acl}

\usepackage{times}
\usepackage{latexsym}
\usepackage[T1]{fontenc}
\usepackage[utf8]{inputenc}
\usepackage{microtype}
\usepackage{inconsolata}

\usepackage{graphicx}
\usepackage{booktabs}
\usepackage{multirow}
\usepackage{amsmath}
\usepackage{amssymb}
\usepackage{xcolor}
\usepackage{enumitem}
\usepackage{url}

\title{What Transfers Under Source Shift? Definitions, Examples, and Fine-Tuning for Climate Disclosure Classification}

\author{
\textbf{Guosheng Li\textsuperscript{1}\textsuperscript{\textdagger}} \quad
\textbf{Fenghui Ren\textsuperscript{1}} \quad
\textbf{Bin Liu\textsuperscript{2}} \quad
\textbf{Chuan Yu\textsuperscript{3}} \\
\textbf{Kaiying Ji\textsuperscript{3}} \quad
\textbf{Lin Yue\textsuperscript{4}} \quad
\textbf{Jun Shen\textsuperscript{1}\textsuperscript{*}} \quad
\textbf{Sasa Qian\textsuperscript{2}} \\
\textsuperscript{1}School of Computing and Information Technology, University of Wollongong \\
\textsuperscript{2}School of Business, University of Wollongong \\
\textsuperscript{3}Business School, University of Sydney \\
\textsuperscript{4}Business School, Macquarie University
}

\begin{document}
\maketitle
\begingroup
\renewcommand{\thefootnote}{\fnsymbol{footnote}}
\footnotetext[1]{Corresponding-author email: \texttt{jshen@uow.edu.au}}
\footnotetext[2]{Email: \texttt{gl934@uowmail.edu.au}}
\endgroup

\begin{abstract}
Climate disclosure classification is a fundamental task for analysing corporate climate disclosures, yet such disclosures appear in many different sources---annual reports, press releases, and earnings calls---that differ in length, purpose, and writing style. Existing evaluations are mostly conducted within a single source, leaving open whether common LLM adaptation strategies remain effective under source shift. We reframe climate disclosure classification as a cross-source adaptation problem and study three widely used adaptation strategies---definitions, examples, and fine-tuning---across eleven open- and closed-source LLMs, using two corpora that share the same label space but come from different sources. We find that all strategies bring positive cross-source gains on average, but the strongest in-source strategies are not the strongest cross-source ones: similarity-based retrieval and LoRA fine-tuning gain most in-source but lose most of that advantage under source shift; randomly selected few-shot examples, a weaker in-source baseline, retain their advantage more reliably; definitions transfer most consistently, though only when their granularity matches the target text. Across these strategies, when the source changes, simpler is often safer. Code and resources are available at \url{https://github.com/Leoccino/TCFD-SourceShift}.
\end{abstract}

\begin{figure}[!b]
\centering
\includegraphics[width=\linewidth]{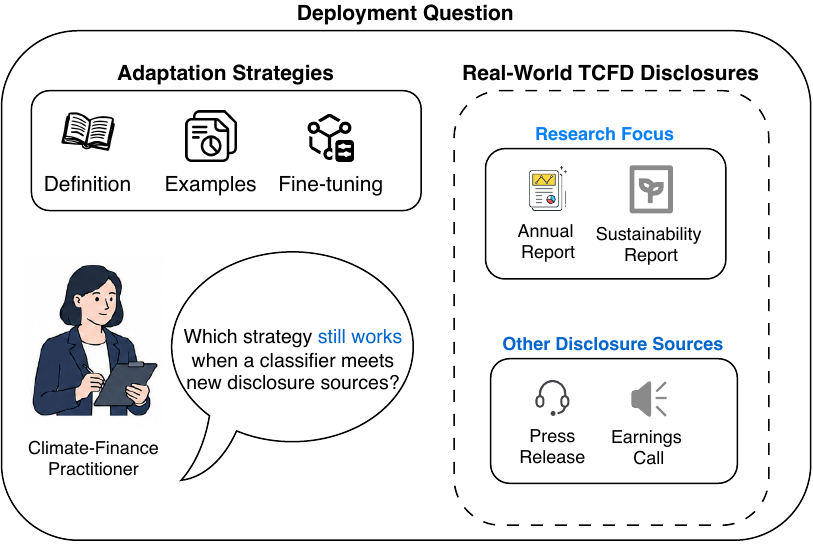}
\caption{Deployment gap motivating this study. Research mainly evaluates TCFD classifiers on annual and sustainability reports, but deployment also requires other disclosure sources. We ask which adaptation strategy still works under this source shift.}
\label{fig:intro-motivation}
\end{figure}

% =================== Section 1: Introduction ===================
\section{Introduction}
\label{sec:intro}

% Climate disclosure classification assigns corporate text to the four pillars of the Task Force on Climate-related Financial Disclosures (TCFD) framework~\citep{fsb2017tcfd,ifrs2023s2}. It is a foundational input for practitioners across climate finance: regulators auditing TCFD reporting quality~\citep{auzepy2023evaluating}, researchers analysing greenwashing~\citep{bingler2022cheap}, and investors constructing firm-level climate-risk indicators from disclosure text~\citep{sautner2023firm}.

Within sustainability-disclosure NLP, we study a climate-focused classification task: assigning corporate text to the four pillars of the Task Force on Climate-related Financial Disclosures (TCFD) framework~\citep{ifrs2023s2}. For climate-finance
practitioners, this task provides a foundation for downstream disclosure
analysis~\citep{auzepy2023evaluating,bingler2022cheap}. Recent work in this area is predominantly LLM-based. Evaluation, however, concentrates on a single class of source: annual reports and corporate sustainability reports~\citep{bingler2022cheap,doi2024automated,calamai2025benchmarking}. In practice, however, TCFD-related information also surfaces in press releases, earnings call transcripts and investor materials---document types that differ from corporate reports in length, purpose, and writing style. The set of sources that current research covers and the set that practitioners actually encounter, in other words, are not the same.

Figure~\ref{fig:intro-motivation} frames this mismatch as a deployment
question. TCFD classification research mainly evaluates report-based sources,
but real-world deployment may involve press releases, earnings calls, and other
disclosure sources. Practitioners therefore need to know whether common
adaptation strategies---richer label definitions, examples, or
fine-tuning---remain reliable
when a classifier meets a new disclosure source. We ask which of these
strategies still transfer under source shift.

We study this question through TCFD disclosure classification, using two
corpora that share the four-class TCFD label space but come from different
disclosure sources. One serves as the adaptation source; the other as a
held-out cross-source target. The three adaptation strategies named in our
title---definitions, examples, and fine-tuning---are swept across eleven
open- and closed-source LLMs and evaluated both in-source and cross-source.

Across these strategies, three findings emerge that diverge from in-source intuition. For definitions, transfer is most consistent when the granularity of the definition matches the granularity of the target text---not when the definition is maximally detailed. For in-context examples, randomly selected few-shot examples often outperform a stronger in-source similarity retriever once the source changes. For fine-tuning, cross-source robustness depends jointly on the learned weights and the inference-time prompt, so a fine-tuned model and its prompt cannot be evaluated in isolation. Across all three, when the source changes, simpler is often safer.

Our contributions are:

\begin{enumerate}

% \item \textbf{A deployment-driven problem formulation.} We study a deployment question faced by climate-finance practitioners: when the disclosure source changes, which adaptation strategy should be chosen? We formulate it as a shared-label cross-source evaluation problem.

\item \textbf{A deployment-oriented source-shift study.} We present, to our
knowledge, the first study of adaptation strategies for TCFD disclosure
classification under source shift. We formulate this deployment question
as a shared-label cross-source evaluation problem.

\item \textbf{Practitioner-oriented evidence for strategy selection.}
Across eleven open- and closed-source LLMs, we examine whether gains from
definitions, examples, and fine-tuning persist. We show that strategies with the strongest in-source gains are not always the safest deployment choices, providing empirical
insights for selecting strategies under source shift.

% \item \textbf{Empirical findings under source shift.} Across eleven open-
% and closed-source LLMs, we compare definitions, examples, and fine-tuning by
% measuring how much of each strategy's in-source gain survives a source shift.
% The results show that single-source rankings can mislead deployment choices,
% giving practitioners a measured basis for strategy selection.

\end{enumerate}

% =================== Section 2: Related Work ===================
\section{Related Work}
\label{sec:related}

\paragraph{Climate disclosure NLP and the single-source gap.}
Climate disclosure NLP has produced domain-specific classifiers around
the TCFD pillars, ranging from BERT-based fine-tuning to LLM prompt
engineering~\citep{webersinke2021climatebert,bingler2022cheap,auzepy2023evaluating,doi2024automated},
with parallel work on related climate-text tasks such as net-zero target
detection, environmental claim detection, and greenwashing-oriented ESG
analysis
\citep{schimanski2023netzero,stammbach2023environmental,ong2025robust}.
Across this literature, evaluation is predominantly conducted within a
single disclosure source, whether by training and testing on the same
corpus or by applying pretrained models to a single report-based
collection~\citep{auzepy2023evaluating,doi2024automated}.
Whether the same models or adaptation strategies remain reliable under
source shift is largely untested. A recent reproducibility
study further finds that many climate-NLP benchmarks rely heavily on
surface-level lexical patterns and contain pervasive annotation issues,
calling into question the reliability of single-source leaderboard
evaluation~\citep{calamai2025benchmarking}.

\paragraph{Cross-source robustness as a framework.}
Outside the climate domain, robustness under distribution shift has been
studied extensively. Classical analysis bounds the out-of-distribution
error in terms of the divergence between source and target
distributions~\citep{ben-david2007analysis}, and large-scale empirical
benchmarks confirm that in-domain performance routinely overestimates
out-of-distribution accuracy even when the label space is held
fixed~\citep{gulrajani2021domainbed,koh2021wilds}. The same pattern
persists in modern NLP: a systematic measurement across thousands of
domain shifts shows that fine-tuned models drop sharply
cross-domain~\citep{calderon2024measuring}, and recent work confirms that
even large language models retain a substantial out-of-domain gap on
text-classification tasks~\citep{roussinov2025controlling}. Our TCFD setting instantiates a broader shared-label cross-source adaptation problem in sustainability-disclosure NLP: the disclosure source varies while the task and label space remain fixed.

\paragraph{Adaptation strategies under source shift.}
Within this framing, the natural question is which of the standard LLM
adaptation mechanisms---task definitions in the prompt, in-context examples,
and parameter-efficient fine-tuning
\citep{brown2020language,dong2024survey,hu2021lora,dettmers2023qlora}---remain
useful under source shift. In-domain studies show that
retrieval-based example selection often improves over random sampling in
in-distribution settings, though the effectiveness of selection strategies
is both data- and
model-dependent~\citep{liu2022makes,peng2024revisiting};
LoRA-style fine-tuning recovers much of the gain of full fine-tuning at a
fraction of the cost. Recent work has begun to compare fine-tuning and
in-context learning side by side under broader distribution-shift
settings~\citep{mosbach2023few}, but whether these
in-domain conclusions transfer when the evaluation source
changes---and which mechanism transfers most reliably---has not been
systematically tested for TCFD or related climate-disclosure tasks. Our
setting is narrower: the task domain and label space remain fixed, while
only the disclosure source changes. We hold the TCFD label space fixed,
vary the disclosure source, and compare definitions, examples, and
fine-tuning under a controlled in-source versus cross-source contrast.

% =================== Section 3: Cross-Source Setting ===================
\section{Cross-Source Setting}
\label{sec:data}

We use two TCFD-labelled corpora that share the same four-class label space
(Governance, Strategy, Risk Management, Metrics \& Targets) but differ in
text style. \textbf{TCFD-CR}~\citep{bingler2022cheap} is drawn from companies'
annual and sustainability reports; we hold out 320 instances as the test
set and split the remaining 1{,}000 into 800 training and 200 validation. \textbf{TCFD-EX}~\citep{rexarski2023tcfd}
is derived from the disclosure examples listed on the TCFD's
example-disclosures list; we use all 593 instances as the cross-source
evaluation set.

\subsection{The two corpora, side by side}
\label{sec:data-examples}

Both items below are labelled \emph{Strategy}:

\begin{quote}\small
\textbf{TCFD-CR:} In October 2017, wildfires started to burn in California
causing damage to many properties [\dots] The fires have forced many people
to be evacuated from their homes [\dots] and have continued to burn in the
early weeks of 2018.\\[3pt]
\textbf{TCFD-EX:} As climate change progresses, the risk of physical damage
and profit loss, such as that from floods and drought, is expected to
increase in the future.
\end{quote}

\begin{figure}[t]
\centering
\includegraphics[width=\linewidth]{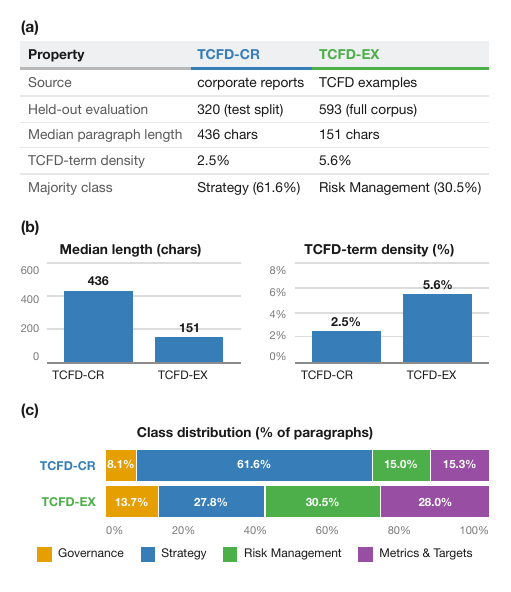}
\caption{TCFD-CR vs.\ TCFD-EX along length, class balance, and lexical style.}
\label{fig:corpora-overview}
\end{figure}

\begin{figure}[t]
\centering
\includegraphics[width=0.95\linewidth]{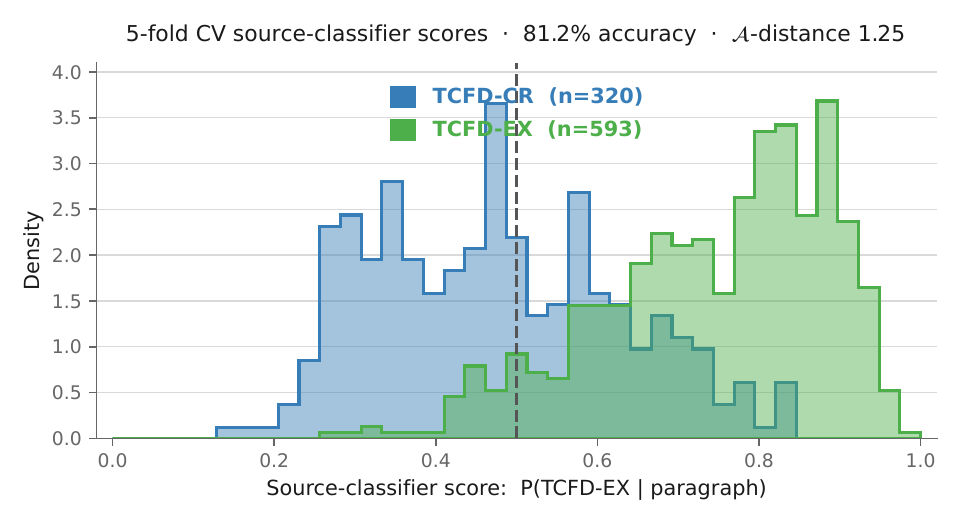}
\caption{A logistic-regression source classifier distinguishes TCFD-CR from
TCFD-EX with 81.2\% 5-fold accuracy.}
\label{fig:source-scores}
\end{figure}

TCFD-CR instances are long and discursive. TCFD-EX instances are short and
direct.
Appendix~\ref{sec:appendix-examples} gives one labelled example pair per class.
Figure~\ref{fig:corpora-overview} quantifies the size, style, and class
contrast between the two corpora. Figure~\ref{fig:source-scores} shows that a
logistic regression on sentence embeddings distinguishes them with 81.2\%
5-fold accuracy, corresponding to a proxy $\mathcal{A}$-distance of
${\approx}1.25$ (out of 2)~\citep{ben-david2007analysis} --- a substantial
source gap in embedding space.

\subsection{Evaluation}
\label{sec:data-directions}

In-source and cross-source evaluation share the same adaptation pool: the
1{,}000 TCFD-CR training and validation instances that feed both examples
and fine-tuning. They differ only in the evaluation set:
\textbf{in-source} uses the TCFD-CR test split (320 instances),
\textbf{cross-source} uses TCFD-EX (593 instances). Fixing the pool isolates
any adaptation that helps in-source but not cross-source as source-specific;
we report macro-F1 throughout.

% =================== Section 4: Experimental Setup ===================
\section{Experimental Setup}
\label{sec:setup}

\begin{figure}[t]
\centering
\includegraphics[width=\linewidth]{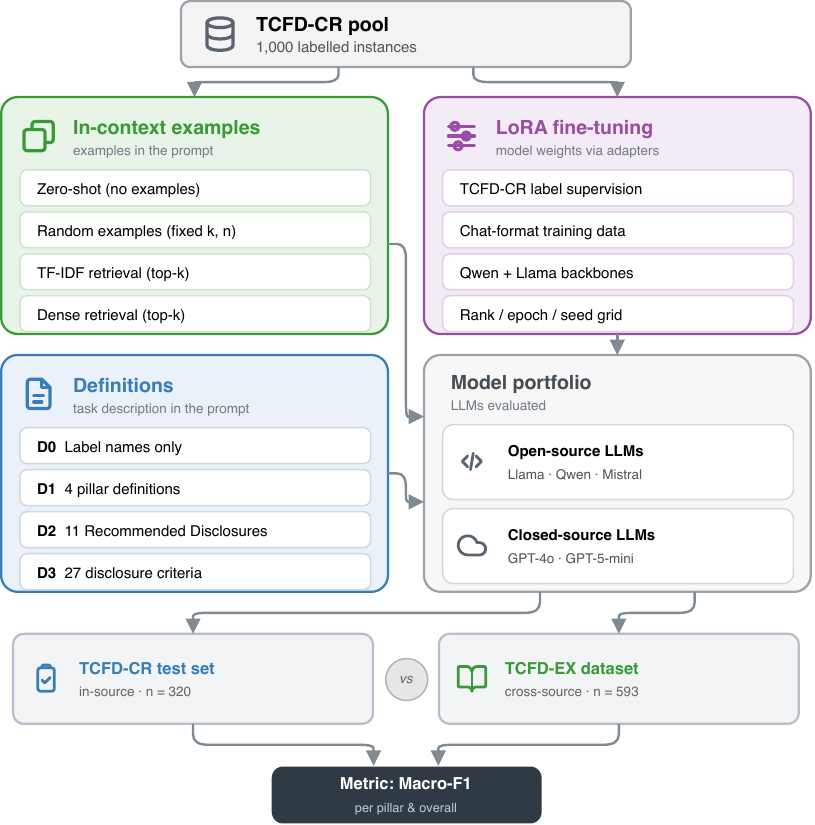}
\caption{Overview of the experimental design: three adaptation strategies ---
definitions, examples, and fine-tuning --- evaluated on both in- and
cross-source datasets.}
\label{fig:setup-overview}
\end{figure}

Figure~\ref{fig:setup-overview} summarises our experimental design. From the
same TCFD-CR adaptation pool, we test three strategies that inject source
signal at different locations: \textbf{definitions} add TCFD label semantics
to the prompt, \textbf{examples} insert labelled instances drawn from the
pool, and \textbf{fine-tuning} updates adapter parameters from source
labels. Each strategy is evaluated across a diverse model
portfolio --- eleven LLMs spanning the Qwen, Llama, and Mistral open-weight
families plus the proprietary GPT-4o and GPT-5-mini --- on both the
in-source TCFD-CR test split ($n{=}320$) and the cross-source TCFD-EX set
($n{=}593$).

\subsection{Adaptation strategies}
\label{sec:setup-adaptation}

\subsubsection{Definition-based adaptation}
\label{sec:setup-definitions}

For definition-based adaptation, we construct a four-level definition
staircase, following the broader use of natural-language label definitions
in zero-shot prompting~\citep{wei2022finetuned,sanh2022multitask}. D0 uses
only the four label names. D1 adds the official descriptions of the four TCFD pillars from the TCFD
Recommendations~\citep{fsb2017tcfd}.
D2 further includes the 11 TCFD Recommended Disclosures. D3 adds 27 criteria distilled from the TCFD implementation
guidance~\citep{doi2024automated}. All definition settings are evaluated in the zero-shot
setting, with no in-context examples, so that performance changes can be
attributed to definition specificity. Appendix~\ref{sec:appendix-prompt-growth} shows a fixed prompt instance under D0--D3, illustrating how only the definition block grows across levels.

\subsubsection{Example-based adaptation}
\label{sec:setup-examples}

For example-based adaptation, we draw examples from the adaptation pool
and vary only the selection rule. We compare
four modes: zero-shot, random examples, TF-IDF similar examples, and dense
similar examples~\citep{liu2022makes, rubin2022learning}. TF-IDF and dense
retrieval cover the two standard notions of textual similarity --- lexical
surface overlap and semantic embedding similarity~\citep{reimers2019sentence}
--- so that the source-shift behaviour of similarity selection is not tied
to a single retriever. The random baseline uses eight fixed examples ($k{=}8,n{=}2$,
two per class). For TF-IDF and dense retrieval, $k$ controls the total number of
retrieved examples included in the prompt, while $n$ caps the maximum number of
examples from any single class to avoid class-dominated prompts. We sweep
$k \in \{3,5,8\}$ and $n \in \{2,3\}$, while fixing the definition level at D1.

\subsubsection{LoRA fine-tuning}
\label{sec:setup-lora}

For parameter-side adaptation, we apply LoRA~\citep{hu2021lora} to four open-source
bases---Llama-3.1-8B~\citep{grattafiori2024llama3}, Qwen3-8B and
Qwen3-32B~\citep{yang2025qwen3}, and
Qwen3.5-9B~\citep{qwen2026qwen35}---spanning two families and the 8--32B range. Adapters are trained on the adaptation pool, using 800 instances
for training and 200 for validation, converted into chat-format
classification examples. Each training
instance uses a minimal label-name prompt, denoted P0, with no pillar
definitions or in-context examples; \S\ref{sec:results-finetuning} reuses this
to test whether supplying definitions at inference time recovers cross-source
performance. The main setting uses 4-bit quantization (QLoRA)~\citep{dettmers2023qlora}, LoRA rank
$r=16$, scaling factor $\alpha=32$, dropout $0.05$, learning rate $2\times10^{-4}$,
an effective batch size of 16, and up to 5 training epochs with best-checkpoint
selection on validation macro-F1. We use three seeds.

\subsection{Models}
\label{sec:setup-models}

We evaluate the prompting strategies on a diverse set of eleven open- and
closed-source LLMs to reduce the risk that our findings are specific to a
single model family or scale. The open-source models cover the
Qwen~\citep{yang2024qwen25,yang2025qwen3,qwen2026qwen35},
Llama~\citep{grattafiori2024llama3,meta2025llama4}, and
Mistral~\citep{mistral2024ministraux} families, with different parameter
sizes and release generations, while GPT-4o~\citep{openai2024gpt4o} and
GPT-5-mini~\citep{openai2025gpt5} are included as closed-source prompted
baselines.
The full model list is reported in Appendix~\ref{sec:appendix-models};
implementation details (serving stack, decoding, retrieval libraries, and LoRA
training) are in Appendix~\ref{sec:appendix-impl}.

% =================== Section 5: Results and Discussion ===================
\section{Results and Discussion}
\label{sec:results}

\subsection{Overview: what transfers under source shift?}
\label{sec:results-overview}

The core question of this paper is what transfers under source shift. Before
turning to the detailed results, we answer it in two steps: first, which
strategies remain effective in the cross-source setting; and second, how much
of their in-source advantage is retained after the source changes.

\begin{figure}[t]
\centering
\includegraphics[width=\linewidth]{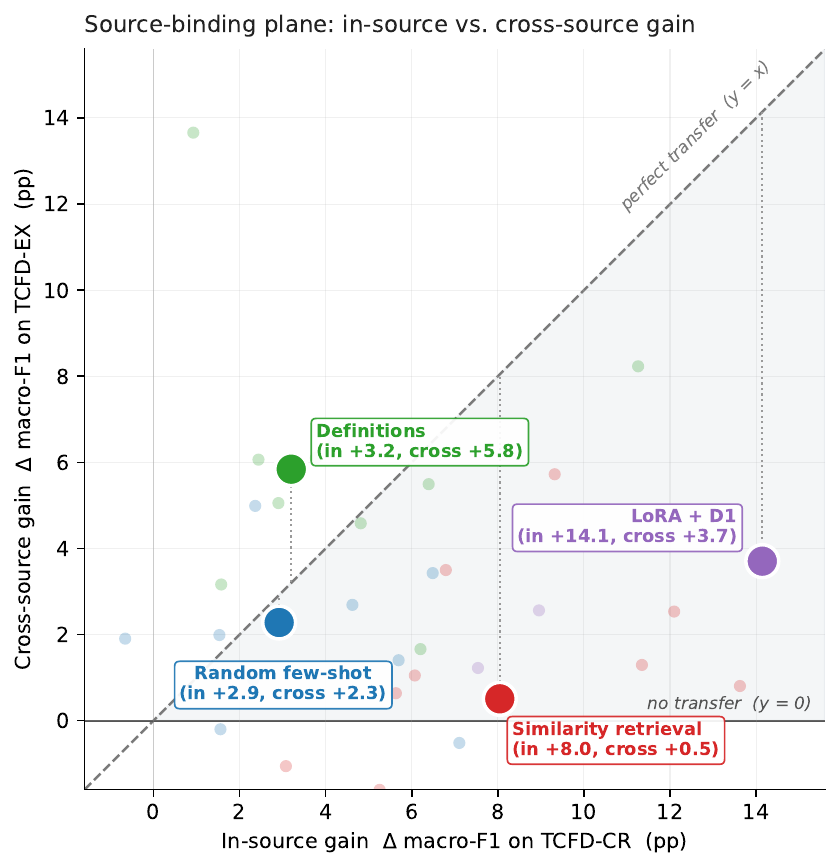}
\caption{Each adaptation strategy as a point in the in-source ($x$) vs.\ cross-source
($y$) $\Delta$ macro-F1 plane; the dashed diagonal $y{=}x$ is perfect transfer.
Faint points are per-model values (per-base for LoRA). Gains are over each
strategy's own no-adaptation baseline (definitions: D2 vs.\ D0; LoRA: adapter
vs.\ base, both at D1; random and similarity: vs.\ D1 zero-shot).}
\label{fig:source-binding-plane}
\end{figure}

\paragraph{Which strategies transfer.}
Figure~\ref{fig:source-binding-plane} shows that all four strategies yield a positive mean cross-source gain. Definitions give the largest improvement, LoRA + D1 and random few-shot show moderate gains, and similarity retrieval barely improves over zero-shot. The per-model points further show that example-based strategies are not always stable: for both random and similarity retrieval, some models fall below zero on TCFD-EX, so introducing TCFD-CR examples can fail to help---or even hurt---on the new source.

\paragraph{How much of the in-source advantage is retained.}
Turning to the second question, a larger in-source gain does not necessarily
survive source shift. Definitions and random few-shot retain their advantage
relatively well. Definitions lie above the dashed diagonal, meaning their
cross-source gain is even larger than their in-source gain; this indicates
that definitions mainly supply TCFD label semantics rather than
source-specific patterns from TCFD-CR. Random few-shot lies close to the
diagonal: although it uses TCFD-CR examples, its mean in-source advantage is
largely retained after the source changes.

By contrast, LoRA + D1 and similarity retrieval fall clearly below the
diagonal, showing that the advantage they gain on TCFD-CR is harder to
transfer. LoRA + D1 yields the largest in-source gain, +14.1 pp, but retains
only +3.7 pp on TCFD-EX, so only part of what fine-tuning learns on TCFD-CR
carries over to the new source. Similarity retrieval is even more
source-bound: it improves TCFD-CR by +8.0 pp but yields almost no gain on
TCFD-EX, only +0.5 pp. The strongest in-source strategy is therefore not
necessarily the strongest cross-source one.

Overall, semantic guidance transfers more readily than advantages tied to
source-specific patterns. When CR and EX exchange roles, retrieval and LoRA
still perform better with same-source adaptation
(Appendix~\ref{sec:appendix-bidirectional}).

\subsection{More specific definitions are not always better}
\label{sec:results-definitions}

\begin{figure}[t]
\centering
\includegraphics[width=\linewidth]{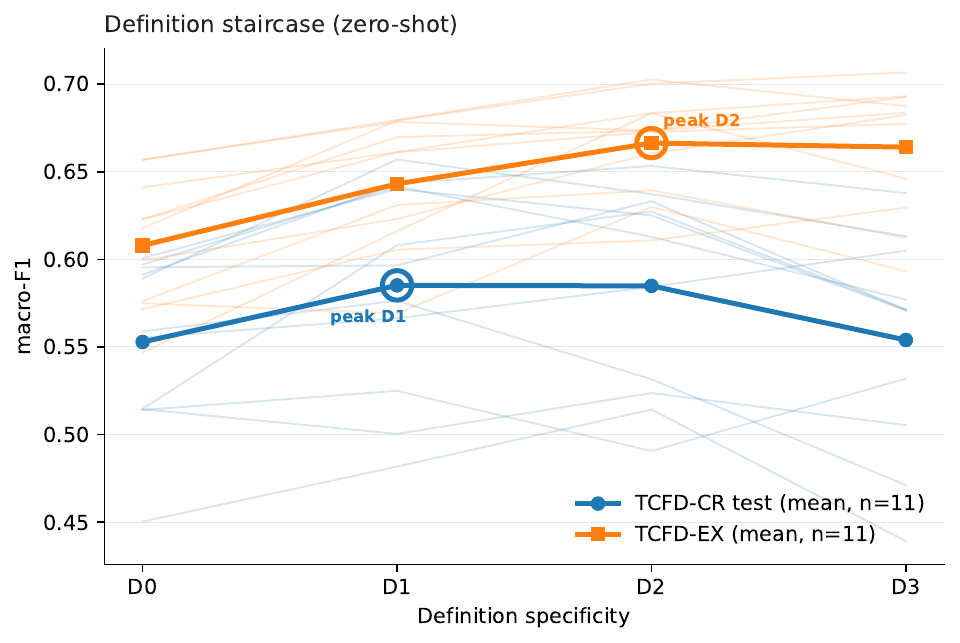}
\caption{Definition specificity under zero-shot prompting, averaged over the
eleven prompted models.}
\label{fig:definitions-mean}
\end{figure}

A natural expectation is that more detailed label definitions should help more:
adding the 11 Recommended Disclosures (D2) or the 27 criteria (D3) on
top of pillar intent (D1) supplies the model with strictly more TCFD knowledge.
Figure~\ref{fig:definitions-mean} shows this expectation breaks at both ends.
Averaged over the eleven prompted models, TCFD-CR peaks at D1 and TCFD-EX
peaks at D2; both sources drop at D3, with the larger drop on TCFD-CR. More
specific definitions are therefore not uniformly better---their effectiveness
depends on whether the granularity matches the target text.

\begin{figure}[t]
\centering
\includegraphics[width=\linewidth]{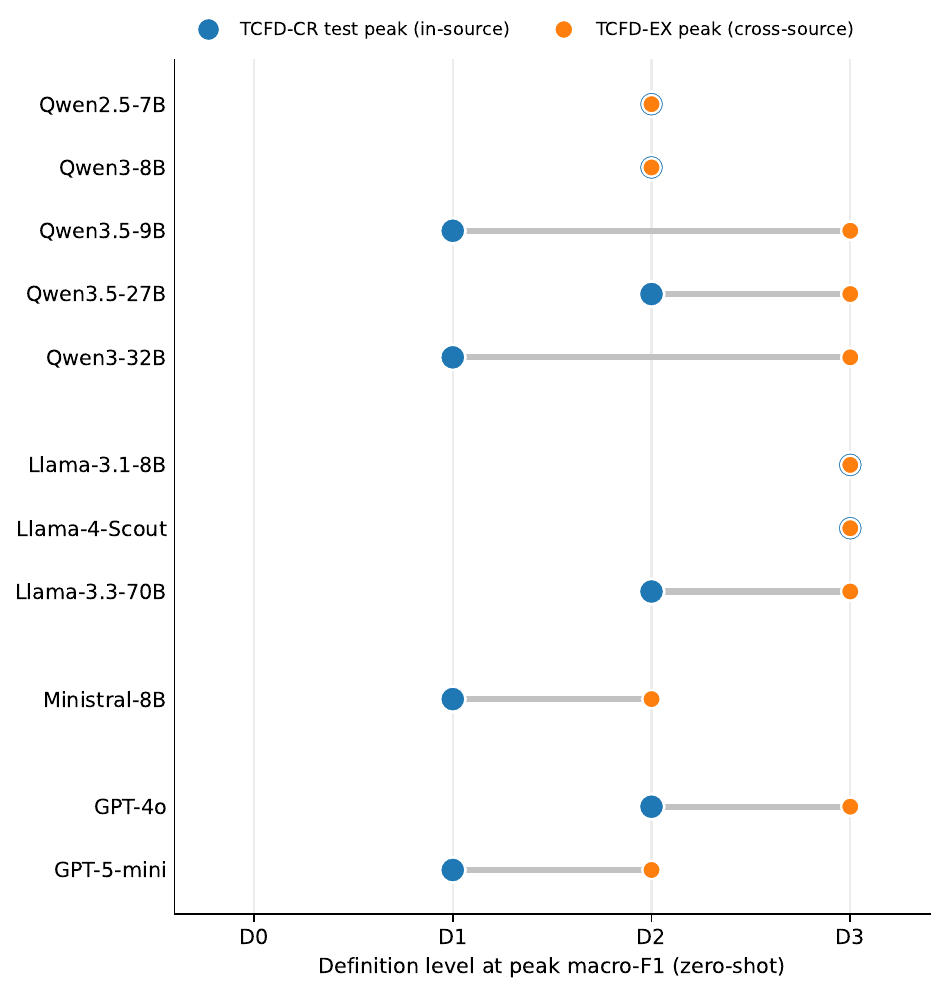}
\caption{Per-model definition peak: the level (D0--D3) at which each model's
zero-shot macro-F1 is highest, on TCFD-CR test (blue) and TCFD-EX (orange).
Models grouped by family, ordered by size.}
\label{fig:definitions-per-model}
\end{figure}

Figure~\ref{fig:definitions-per-model} confirms this at the model level: peaks scatter across D1--D3 on TCFD-CR but cluster at D2--D3 on TCFD-EX. Our claim is therefore not that one definition level is always optimal. The
stable pattern is directional: no model peaks earlier on TCFD-EX than on
TCFD-CR. In other words, the TCFD-EX target tends to benefit from more specific
definitions. Full per-model macro-F1 across D0--D3 is in
Appendix~\ref{sec:appendix-defs-perm}.

This pattern is consistent with the text characteristics of the two sources.
TCFD-CR comes from real corporate reports, where instances are longer and more
likely to mix business context, climate actions, financial figures, and
target-related statements. More specific definitions may therefore push the
model toward local keywords and away from the instance's overall meaning. By
contrast, TCFD-EX instances are shorter, more focused, and closer to the
wording of the TCFD Recommended Disclosures. As a
result, D2 better matches the granularity of TCFD-EX. Appendix~\ref{sec:appendix-def-cases} provides two Qwen3-32B zero-shot cases
illustrating this peak-shift mechanism: D1 works better for a noisy TCFD-CR
instance, while D2 works better for a TCFD-EX instance whose wording closely
matches a Recommended Disclosure.

To test whether definitions also benefit from matching the target source, we
constructed CR-specific definitions of the same length as D2; they outperformed
D2 on CR but not EX (Appendix~\ref{sec:appendix-crsem20}).

\paragraph{Takeaway.}
Definitions transfer because they encode TCFD label semantics rather than
source-specific examples. However, more detail is not always better. Together,
these results suggest that the best definition level depends on the target
source: definitions appear to help most when their granularity matches the
target text.

\subsection{Better in-source retrieval does not imply better transfer}
\label{sec:results-examples}
A natural expectation is that more refined example selection should help more: retrieving examples similar to the target instance should outperform random sampling. Table~\ref{tab:examples-landscape} reports the full per-model
example-selection results. To better match a realistic deployment setting, for
TF-IDF and dense retrieval we use, for each model, the $(k,n)$ configuration
that performs best on TCFD-CR, and apply that same configuration directly to
both TCFD-CR and TCFD-EX. Under this setting, random few-shot improves macro-F1
by +2.9 pp on TCFD-CR and by +2.3 pp on TCFD-EX. By contrast, TF-IDF and dense
retrieval improve TCFD-CR by +8.0 pp and +7.9 pp respectively, but improve
TCFD-EX by only +0.5 pp and +1.0 pp. Thus, retrieval is highly effective when
the target texts and the adaptation pool come from the same source, but this
advantage does not transfer well to a new source.

\begin{table*}[t]
\centering\small
\setlength{\tabcolsep}{5pt}
\renewcommand{\arraystretch}{1.05}
\begin{tabular}{@{}l cc cc cc cc@{}}
\toprule
& \multicolumn{2}{c}{\textbf{Zero-shot}}
& \multicolumn{2}{c}{\textbf{Random $k{=}8,n{=}2$}}
& \multicolumn{2}{c}{\textbf{TF-IDF best-on-CR}}
& \multicolumn{2}{c}{\textbf{Dense best-on-CR}} \\
\cmidrule(lr){2-3} \cmidrule(lr){4-5} \cmidrule(lr){6-7} \cmidrule(lr){8-9}
\textbf{Model} & CR & EX & CR & EX & CR & EX & CR & EX \\
\midrule
Qwen2.5-7B & 0.482 & 0.569 & 0.524 & \textbf{0.634} & \textbf{0.575} & 0.626 & 0.563 & 0.580 \\
Qwen3-8B & 0.500 & 0.616 & 0.557 & 0.630 & 0.614 & 0.629 & \textbf{0.625} & \textbf{0.658} \\
Qwen3.5-9B & 0.642 & 0.661 & 0.665 & 0.671 & 0.702 & \textbf{0.672} & \textbf{0.714} & 0.667 \\
Qwen3.5-27B & 0.643 & 0.678 & 0.689 & 0.705 & \textbf{0.711} & \textbf{0.713} & 0.708 & 0.689 \\
Qwen3-32B & 0.640 & 0.661 & 0.655 & \textbf{0.681} & 0.696 & 0.668 & \textbf{0.713} & 0.668 \\
Llama-3.1-8B & 0.525 & 0.605 & 0.549 & \textbf{0.655} & \textbf{0.661} & 0.614 & 0.632 & 0.639 \\
Llama-4-Scout & 0.566 & \textbf{0.670} & 0.638 & 0.665 & 0.681 & 0.630 & \textbf{0.706} & 0.641 \\
Llama-3.3-70B & 0.597 & 0.623 & 0.661 & 0.657 & \textbf{0.718} & 0.648 & 0.696 & \textbf{0.671} \\
Ministral-8B & 0.576 & 0.631 & 0.545 & \textbf{0.650} & \textbf{0.615} & 0.598 & 0.586 & 0.625 \\
GPT-4o & 0.608 & \textbf{0.679} & 0.624 & 0.677 & 0.661 & 0.663 & \textbf{0.673} & 0.670 \\
GPT-5-mini & 0.657 & 0.679 & 0.650 & \textbf{0.699} & 0.688 & 0.669 & \textbf{0.692} & 0.674 \\
\midrule
\textit{Mean} & 0.585 & 0.643 & 0.614 & \textbf{0.666} & \textbf{0.666} & 0.648 & 0.664 & 0.653 \\
\midrule
\textit{Mean $\Delta$ vs zero-shot (pp)} & \multicolumn{2}{c}{---} & +2.9 & +2.3 & +8.0 & +0.5 & +7.9 & +1.0 \\
\bottomrule
\end{tabular}
\caption{Macro-F1 on TCFD-CR (in-source) and TCFD-EX (cross-source) for four
selection rules. Random uses fixed $k{=}8,n{=}2$; TF-IDF and Dense use the
per-model best-on-CR $(k,n)$, applied to both settings. \textbf{Bold} marks the
best value per model in each setting; last row is the mean $\Delta$ over
zero-shot.}
\label{tab:examples-landscape}
\end{table*}

\begin{figure}[t]\centering\includegraphics[width=\linewidth]{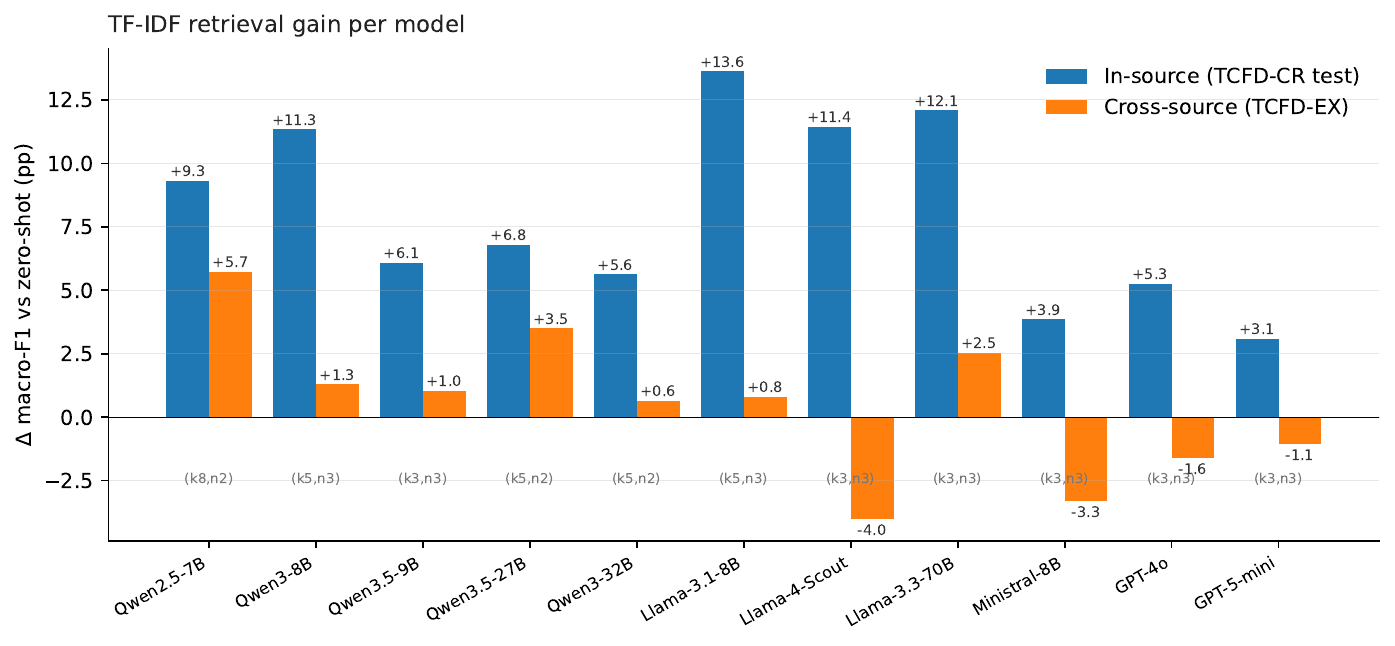}
\caption{TF-IDF retrieval gain over zero-shot, per model, with $(k,n)$ tuned
per model on TCFD-CR. Blue: in-source (TCFD-CR test); orange: cross-source
(TCFD-EX).}
\label{fig:retrieval}
\end{figure}

Figure~\ref{fig:retrieval} shows that this finding is not an artifact of
averaging. For all prompted models, TF-IDF retrieval's gain on TCFD-CR exceeds
its gain on TCFD-EX, and for some models performance on TCFD-EX even drops. The
cross-source problem is therefore not only that the gain becomes smaller: for
some models, similarity retrieval directly hurts performance under source
shift.

\begin{figure}[t]
\centering
\includegraphics[width=\linewidth]{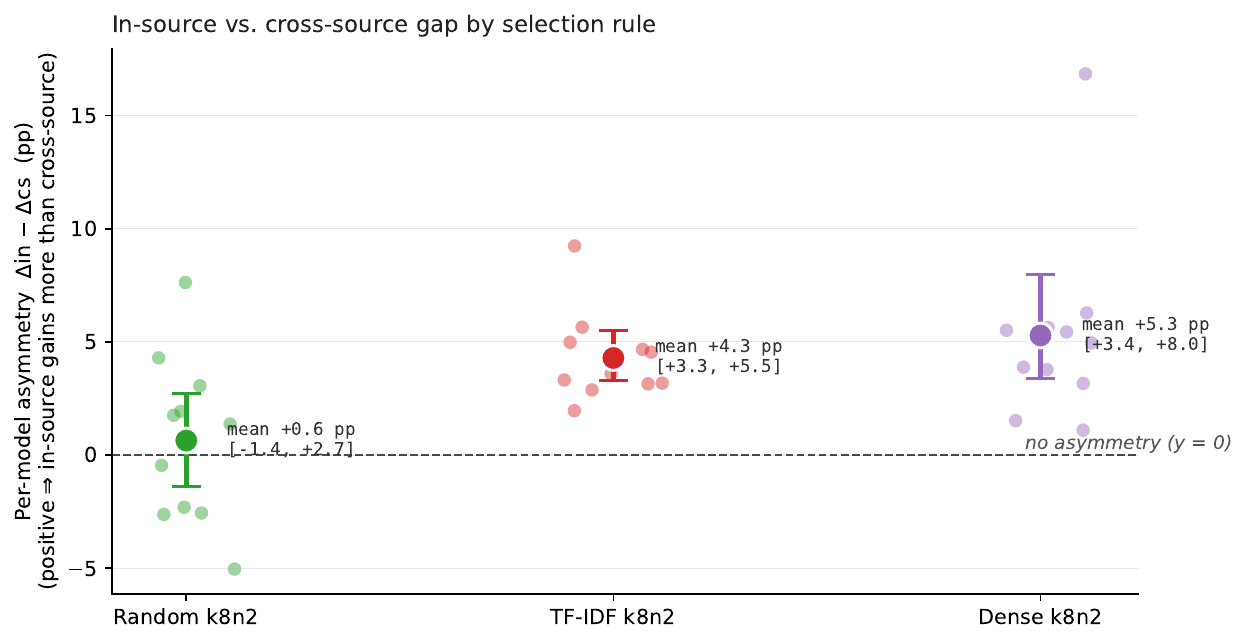}
\caption{Per-model in-source minus cross-source gap for each selection rule,
all fixed at $k{=}8$, $n{=}2$ on the same TCFD-CR adaptation pool. Light dots are
per-model; the point and bar mark the mean and its 95\% paired-bootstrap CI.}
\label{fig:retrieval-isolation}
\end{figure}

The results above reflect a realistic best-on-CR setting, but they do not yet
fully isolate the source of the cross-source drop, because different methods
and different models may use different $(k,n)$ configurations. To separate
these factors, Figure~\ref{fig:retrieval-isolation} fixes the prompt, the
adaptation pool, and the example budget, and varies only the example selection
rule. In this aligned comparison, random few-shot shows no clearly stable
in-source--cross-source gap, whereas TF-IDF and dense retrieval show a larger
gap. This indicates that the source-bound effect comes mainly from the
per-target similarity-selection step.

With two examples per label, similar examples from the correct class help less
when sources differ, while similar examples from other classes remain similarly
harmful (Appendix~\ref{sec:appendix-retrieval-mechanism}).

\paragraph{Takeaway.}
Few-shot examples transfer across sources, but more refined selection does not
necessarily transfer better. On TCFD-CR, similarity retrieval beats random by a
wide margin; on TCFD-EX, random is the more reliable choice. The failure point
is the assumption behind retrieval, namely that textual similarity approximates
label similarity. This assumption is useful when the target text and adaptation
pool share a source, but weakens under source shift. The same asymmetry holds
across the full $k\times n$ grid for both retrievers (${\sim}5$--$7$\,pp
in-source vs.\ ${\le}1$\,pp cross-source; Appendix~\ref{sec:appendix-knsweep}).
For cross-source few-shot adaptation, a simple random baseline is therefore a
stronger default than a stronger retriever.

\subsection{Fine-tuning is in-source champion but not cross-source champion}
\label{sec:results-finetuning}

\begin{figure*}[t]
\centering
\includegraphics[width=0.92\textwidth]{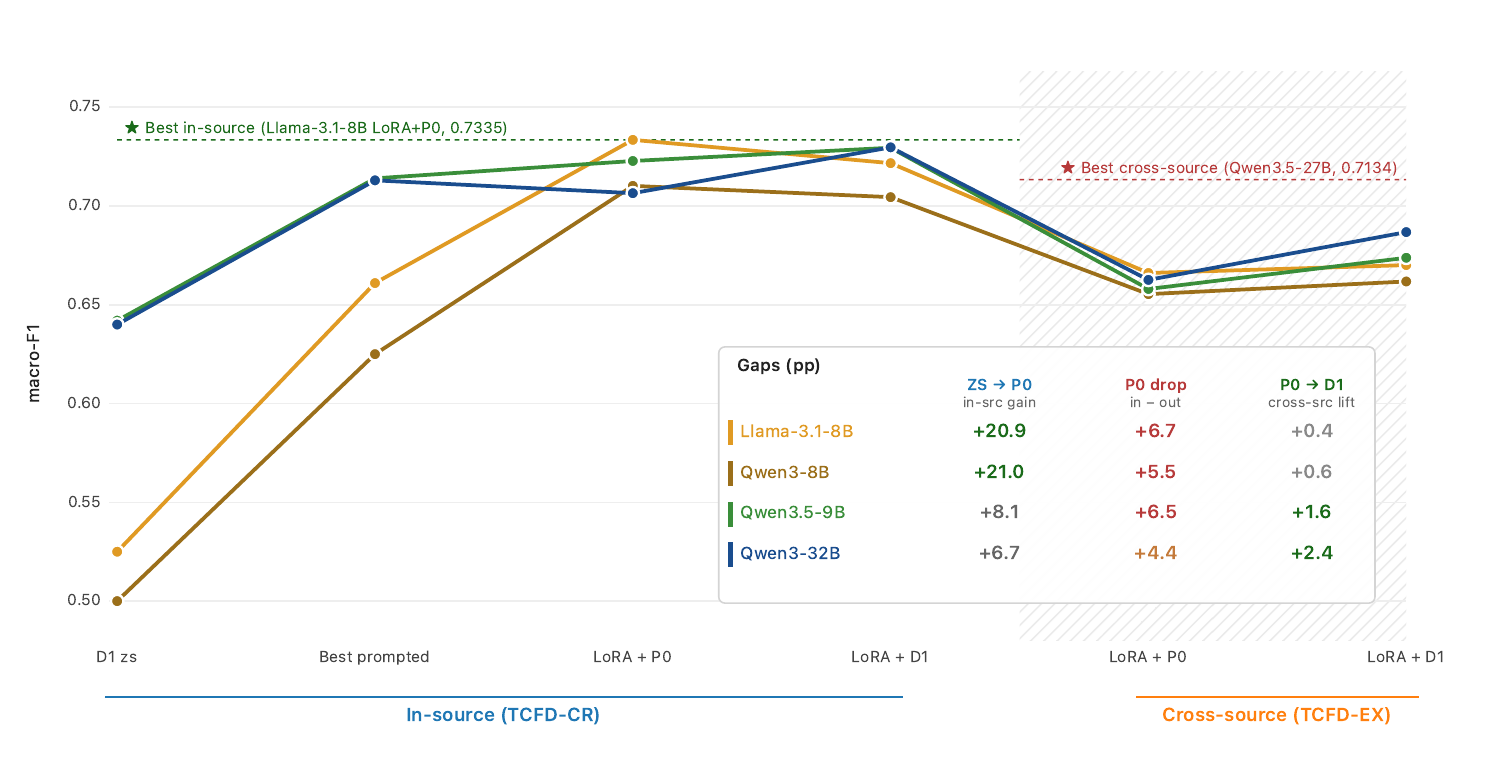}
\caption{Per-base LoRA trajectory across in-source (TCFD-CR) and cross-source
(TCFD-EX) evaluation, with two non-LoRA baselines on the left. All LoRA values
are three-seed means.}
\label{fig:lora-transfer-diagnosis}
\end{figure*}

Compared with prompt engineering, fine-tuning is a stronger form of adaptation
because it updates model parameters directly from labelled supervision. In our
setting, this leads to the best in-source performance on TCFD-CR. The key
question is whether this parameter-side learning captures transferable TCFD
label semantics, or mainly fits source-specific decision patterns from TCFD-CR.

Figure~\ref{fig:lora-transfer-diagnosis} shows
that LoRA gives a large in-source gain on TCFD-CR for every fine-tuned
base. LoRA also outperforms the best prompted
strategy for two of the four bases; for Qwen3.5-9B and Qwen3-32B the best prompted
strategy is slightly higher. This confirms that supervised fine-tuning is highly
effective at fitting the TCFD-CR source distribution.

However, this source fit does not fully transfer. Figure~\ref{fig:lora-transfer-diagnosis} also shows that
the same LoRA adapters drop when evaluated on TCFD-EX with the training prompt (P0). The cross-source drop is consistent across all four
bases, ranging from $-4.4$\,pp to $-6.7$\,pp. This indicates that LoRA
learns useful decision patterns for TCFD-CR, but part of this advantage is tied
to the source.

Because LoRA training updates the model parameters directly from text--label pairs alone, the adapters never see any TCFD definition; all label semantics must be inferred from the supervision signal itself. This motivates a second test: if those semantics are not in the adapters, can we supply them at inference time instead? Keeping the trained adapters fixed, we replace the training-time prompt P0 with the D1 definition prompt and re-evaluate. Figure~\ref{fig:lora-transfer-diagnosis} shows that D1 improves cross-source performance for all four bases, with gains from +0.4 to +2.4\,pp. Its in-source effect, however, is mixed: D1 slightly reduces performance on Llama-3.1-8B and Qwen3-8B, but improves Qwen3.5-9B and Qwen3-32B. D1 therefore does not make LoRA uniformly better; rather, it recovers part of the transferable label semantics under source shift, with larger gains on the stronger base models. Per-base numbers are in Appendix~\ref{sec:appendix-lora-perfam}; a cross-strategy summary, including the supervised ClimateBERT-TCFD reference, is in Appendix~\ref{sec:appendix-best-of}.

\paragraph{Takeaway.}

LoRA yields the largest in-source gain on TCFD-CR, yet all four fine-tuned
bases drop under source shift: the in-source champion is not the cross-source
one. Since LoRA never sees explicit TCFD definitions during training, we keep
the adapters fixed and add D1 definitions at inference time. Cross-source
performance consistently recovers, indicating that part of the loss reflects
label semantics missing from the training signal, rather than something that
can be solved by the learned weights alone. Cross-source robustness for
fine-tuned models should therefore be treated as a joint design problem over
weights and prompts, with the prompt-side correction larger on stronger base
models.

% =================== Section 6: Conclusion ===================
\section{Conclusion}
\label{sec:conclusion}

We studied TCFD classification as a cross-source adaptation problem, using
TCFD-CR as source and TCFD-EX as the target. Across definitions,
examples, and fine-tuning, stronger in-source adaptation strategy does
not necessarily imply better transfer.

Definitions transfer most reliably, but their granularity must match the target
text. Examples can transfer, but similarity retrieval is source-bound: a strong
in-source retriever can be less reliable than random examples under source
shift. Fine-tuning gives the best in-source performance, yet only partly
transfers; adding definition at inference time partly recovers the
cross-source loss without retraining.

Practitioners may naturally choose the strategy that performs best on the
available source, but our results show that this can be unsafe under
source shift. We therefore translate the findings into safer defaults:
match the definition prompt to the target text, prefer a small random
few-shot pool over a similarity retriever, and keep a definition prompt at
inference time even for fine-tuned models. The strongest in-source choice
is rarely the strongest cross-source one.

% =================== Limitations (unnumbered, ACL/ARR style) ===================
\section*{Limitations}
\label{sec:limitations}

Our cross-source evaluation is built on a single dataset pair, adapting on
TCFD-CR and evaluating on TCFD-EX. To our knowledge, these are currently the
only publicly available datasets that share the TCFD four-pillar label space
across sources, which constrains cross-source evaluation to this pair. The two
corpora nonetheless exhibit a non-trivial distributional gap (proxy
$\mathcal{A}$-distance 1.25; 81.2\% 5-fold source-classifier accuracy;
\S\ref{sec:data-examples}), suggesting that this is not a near-duplicate
evaluation but a meaningful setting for practitioners who need to anticipate
how adaptation strategies behave under source shift.

\bibliography{custom}

\appendix
% =================== Appendix ===================

\section{Reverse-direction control}
\label{sec:appendix-bidirectional}

To test whether the result depends on the original CR-to-EX direction, we
reversed the adaptation direction while keeping each evaluation target fixed.
Retrieval results average Qwen3-8B, Llama-3.1-8B, Qwen3.5-9B, and Qwen3-32B;
LoRA + D1 results average Qwen3-8B and Qwen3.5-9B. CR uses 800 adaptation
and 320 test texts. EX uses 351 adaptation and 123 test texts for retrieval;
LoRA uses 116 EX test texts after excluding seven that overlap its training or
validation data.

\begin{center}
\scriptsize
\setlength{\tabcolsep}{6pt}
\renewcommand{\arraystretch}{1.05}
\begin{tabular}{@{}llcc@{}}
\toprule
\textbf{Method} & \textbf{Target} & \textbf{Same-source} & \textbf{Cross-source} \\
\midrule
TF-IDF & CR & \textbf{0.6385} & 0.5872 \\
Dense & CR & \textbf{0.6507} & 0.5804 \\
TF-IDF & EX & \textbf{0.7669} & 0.7059 \\
Dense & EX & \textbf{0.7490} & 0.7317 \\
\midrule
LoRA + D1 & CR & \textbf{0.7170} & 0.6152 \\
LoRA + D1 & EX & \textbf{0.7490} & 0.6986 \\
\bottomrule
\end{tabular}
\captionof{table}{Reverse-direction control (macro-F1).}
\label{tab:bidirectional}
\end{center}

Same-source adaptation is higher in all six comparisons. These controls use
smaller model subsets than the main study and therefore confirm the direction
of the result rather than replace the primary analysis.

\section{Per-class examples from each corpus}
\label{sec:appendix-examples}

Table~\ref{tab:examples} shows one example per class from TCFD-CR and TCFD-EX,
extending the single Strategy pair in \S\ref{sec:data-examples}.

\begin{table*}[!tb]
\centering
\small
\setlength{\tabcolsep}{3pt}
\begin{tabular}{@{}p{0.10\textwidth} p{0.405\textwidth} p{0.405\textwidth}@{}}
\toprule
Class & \textbf{TCFD-CR} (corporate report) & \textbf{TCFD-EX} (TCFD example) \\
\midrule
Governance & Report 2019 (SR 2019), as requested by the shareholders. The hardcopies will be delivered once they are made available to the Company. [\dots] The environmental concerns like global warming, deforestation, climate change [\dots] affect every human, animal and nation on this planet. & The ARCGS oversees the implications of sustainable development issues for the company under five sustainability themes, of which one is climate change. \\
\addlinespace
Strategy & In October 2017, wildfires started to burn in California causing damage to many properties [\dots] The fires have forced many people to be evacuated from their homes and neighbourhoods and have continued to burn in the early weeks of 2018. & As climate change progresses, the risk of physical damage and profit loss, such as that from floods and drought, is expected to increase in the future. \\
\addlinespace
Risk Mgmt. & In our view, true sustainable investing cannot be achieved by simply voting a proxy, adding a director of sustainability or even divesting from an asset class. [\dots] It requires integration across our products, across our product teams and across our entire organization. & Our focus for emerging risks is on reducing the impact should an event occur, and on advocacy efforts to reduce the likelihood of the risk manifesting. \\
\addlinespace
Metrics \& Targets & During 2019, the Group produced approx.\ 14{,}059~MWh of electrical energy from renewable sources, 38\% of which used for self-consumption on site [\dots] 90\% of production comes from renewable sources installed in Italy [\dots] & Note 1 Targets are from a baseline of 12 months to 31 December 2020: Power emissions intensity --- 321~KgCO\textsubscript{2}/MWh; Energy absolute emissions --- 75.0~MtCO\textsubscript{2}. \\
\bottomrule
\end{tabular}
\caption{One example per class from each corpus. TCFD-CR is longer, narrative,
and mixes climate content with business context; TCFD-EX is short declarative
examples. Both items in a row carry the same class label.}
\label{tab:examples}
\end{table*}

\section{Prompt growth under the definition staircase}
\label{sec:appendix-prompt-growth}

Figure~\ref{fig:prompt-growth} shows one fixed classification instance under
the four definition levels. The task instruction, target instance, and output
constraint are unchanged; only the definition block grows from label names
(D0) to pillar intent (D1), Recommended Disclosures (D2), and criteria
(D3).

\begin{figure*}[!tb]
\centering
\small

\fcolorbox{black!30}{black!3}{
\begin{minipage}{0.94\textwidth}
\textbf{Fixed prompt frame used in all four settings}

\medskip
\texttt{Task: Classify the following text into one of the TCFD-recommended
disclosure categories --- the one whose pillar intent the instance best matches.}

\medskip
\texttt{[Definition block inserted here: D0 / D1 / D2 / D3]}

\medskip
\texttt{Paragraph to classify:}

\textit{Since 2016 we have been testing the resilience of our portfolio against
the scenarios from the IEAs World Energy Outlook (WEO) report.}

\medskip
\texttt{Respond with EXACTLY one of these four labels and nothing else:
Governance, Strategy, Risk Management, Metrics \& Targets.}
\end{minipage}
}

\vspace{0.8em}

\begin{tabular}{@{}p{0.12\textwidth} p{0.80\textwidth}@{}}
\toprule
\textbf{Level} & \textbf{Definition block inserted into the same prompt} \\
\midrule

\textbf{D0}
\newline
{\scriptsize label names}
&
\fcolorbox{black!20}{white}{
\begin{minipage}{0.77\textwidth}
Categories:
\begin{itemize}[leftmargin=1.2em,nosep]
\item Governance
\item Strategy
\item Risk Management
\item Metrics \& Targets
\end{itemize}
\end{minipage}
}
\\
\addlinespace[0.8em]

\textbf{D1}
\newline
{\scriptsize + pillar intent}
&
\fcolorbox{black!20}{blue!3}{
\begin{minipage}{0.77\textwidth}
Categories and Definitions:

\medskip
\textbf{Governance:} Disclose the organization's governance around
climate-related risks and opportunities.

\textbf{Strategy:} Disclose the actual and potential impacts of
climate-related risks and opportunities on the organization's businesses,
strategy, and financial planning where such information is material.

\textbf{Risk Management:} Disclose how the organization identifies, assesses,
and manages climate-related risks.

\textbf{Metrics \& Targets:} Disclose the metrics and targets used to assess
and manage relevant climate-related risks and opportunities where such
information is material.
\end{minipage}
}
\\
\addlinespace[0.8em]

\textbf{D2}
\newline
{\scriptsize + 11 RDs}
&
\fcolorbox{black!20}{green!3}{
\begin{minipage}{0.77\textwidth}
D1, plus the official Recommended Disclosures under each pillar.

\medskip
For example, under \textbf{Strategy}:
\begin{itemize}[leftmargin=1.2em,nosep]
\item Describe the climate-related risks and opportunities identified over the
short, medium, and long term.
\item Describe the impact of climate-related risks and opportunities on
businesses, strategy, and financial planning.
\item Describe the resilience of the organization's strategy, taking into
consideration different climate-related scenarios, including a $2^\circ$C or
lower scenario.
\end{itemize}
\end{minipage}
}
\\
\addlinespace[0.8em]

\textbf{D3}
\newline
{\scriptsize + 27 criteria}
&
\fcolorbox{black!20}{orange!5}{
\begin{minipage}{0.77\textwidth}
D2, plus criteria distilled from the TCFD implementation guidance.

\medskip
For example, under \textbf{Strategy}:
\begin{itemize}[leftmargin=1.2em,nosep]
\item 03-01: Time horizon(s) for climate-related risks and opportunities.
\item 04-01: Impact on businesses or strategy.
\item 04-02: Impact on financial planning.
\item 05-02: Climate-related scenarios and associated time horizon(s).
\item 05-03: Impact of climate-related scenarios on the strategy.
\item 05-05: Impact of climate-related scenarios on financial planning.
\end{itemize}

Across all pillars, D3 contains 27 criteria.
\end{minipage}
}
\\

\bottomrule
\end{tabular}

\caption{A single prompt instance under the definition staircase. The frame,
target instance, and output constraint are fixed; only the inserted definition
block grows from D0 to D3.}
\label{fig:prompt-growth}
\end{figure*}

\section{Model portfolio}
\label{sec:appendix-models}

Table~\ref{tab:models} summarises the model portfolio used in the prompting
experiments.

\begin{table*}[!tb]
\centering\small
\renewcommand{\arraystretch}{1.1}
\begin{tabular}{@{}lrl@{}}
\toprule
Model & Size & Release \\
\midrule
Qwen2.5-7B-Instruct        & 7B  & Sep 2024 \\
Qwen3-8B-FP8               & 8B  & Apr 2025 \\
Qwen3.5-9B                 & 9B  & Feb 2026 \\
Qwen3.5-27B-FP8            & 27B & Feb 2026 \\
Qwen3-32B                  & 32B & Apr 2025 \\
\midrule
Llama-3.1-8B-Instruct      & 8B  & Jul 2024 \\
Llama-4-Scout              & 17B active & Apr 2025 \\
Llama-3.3-70B-Instruct     & 70B & Dec 2024 \\
\midrule
Ministral-8B-Instruct-2410 & 8B  & Oct 2024 \\
\midrule
GPT-4o                     & --  & May 2024 \\
GPT-5-mini                 & --  & Aug 2025 \\
\bottomrule
\end{tabular}
\caption{Model portfolio used in the prompting experiments.}
\label{tab:models}
\end{table*}

\section{Implementation details}
\label{sec:appendix-impl}

Open-weight LLMs are served with vLLM~0.19.1~\citep{kwon2023efficient} on
$4\times$ NVIDIA A100 40\,GB GPUs. GPT-4o and GPT-5-mini are queried through the OpenAI Chat Completions
API with temperature set to 0. Outputs are parsed against a strict four-label
format; outputs that fail to parse are marked as INVALID. TF-IDF retrieval is
implemented with scikit-learn defaults, and dense retrieval uses
\texttt{all-mpnet-base-v2} sentence embeddings. LoRA fine-tuning uses
PEFT~\citep{mangrulkar2022peft} with 4-bit loading, BF16 compute, and
gradient checkpointing.

\section{Per-model macro-F1 across the definition staircase}
\label{sec:appendix-defs-perm}

Table~\ref{tab:appendix-defs-perm} reports the full per-model numbers behind
Figures~\ref{fig:definitions-mean} and~\ref{fig:definitions-per-model}.

\begin{table*}[!tb]
\centering\small
\setlength{\tabcolsep}{4pt}
\renewcommand{\arraystretch}{1.05}
\begin{tabular}{@{}l cc cc cc cc@{}}
\toprule
 & \multicolumn{2}{c}{\textbf{D0}}
 & \multicolumn{2}{c}{\textbf{D1}}
 & \multicolumn{2}{c}{\textbf{D2}}
 & \multicolumn{2}{c}{\textbf{D3}} \\
\cmidrule(lr){2-3} \cmidrule(lr){4-5} \cmidrule(lr){6-7} \cmidrule(lr){8-9}
\textbf{Model} & CR & EX & CR & EX & CR & EX & CR & EX \\
\midrule
Qwen2.5-7B & 0.450 & 0.575 & 0.482 & 0.569 & \textbf{0.514} & \textbf{0.630} & 0.439 & 0.593 \\
Qwen3-8B & 0.514 & 0.547 & 0.500 & 0.616 & \textbf{0.524} & \textbf{0.683} & 0.505 & 0.646 \\
Qwen3.5-9B & 0.597 & 0.641 & \textbf{0.642} & 0.661 & 0.613 & 0.673 & 0.577 & \textbf{0.677} \\
Qwen3.5-27B & 0.591 & 0.657 & 0.643 & 0.678 & \textbf{0.653} & 0.674 & 0.638 & \textbf{0.693} \\
Qwen3-32B & 0.601 & 0.623 & \textbf{0.640} & 0.661 & 0.625 & 0.683 & 0.571 & \textbf{0.693} \\
Llama-3.1-8B & 0.514 & 0.572 & 0.525 & 0.605 & 0.491 & 0.611 & \textbf{0.532} & \textbf{0.629} \\
Llama-4-Scout & 0.555 & 0.623 & 0.566 & 0.670 & 0.584 & 0.673 & \textbf{0.605} & \textbf{0.684} \\
Llama-3.3-70B & 0.595 & 0.599 & 0.597 & 0.623 & \textbf{0.633} & 0.660 & 0.571 & \textbf{0.682} \\
Ministral-8B & 0.559 & 0.576 & \textbf{0.576} & 0.631 & 0.532 & \textbf{0.639} & 0.471 & 0.612 \\
GPT-4o & 0.515 & 0.618 & 0.608 & 0.679 & \textbf{0.627} & 0.700 & 0.571 & \textbf{0.707} \\
GPT-5-mini & 0.589 & 0.657 & \textbf{0.657} & 0.679 & 0.637 & \textbf{0.703} & 0.613 & 0.687 \\
\midrule
\textit{Mean} & 0.553 & 0.608 & \textbf{0.585} & 0.643 & 0.585 & \textbf{0.666} & 0.554 & 0.664 \\
\bottomrule
\end{tabular}
\caption{Per-model zero-shot macro-F1 across the four definition levels, on
TCFD-CR test (in-source) and TCFD-EX (cross-source). \textbf{Bold} marks each
row's best level per setting; the mean row averages over the eleven prompted
models.}
\label{tab:appendix-defs-perm}
\end{table*}

\section{Case examples for the definition peak shift}
\label{sec:appendix-def-cases}

Figure~\ref{fig:def-cases} shows two Qwen3-32B zero-shot predictions on
Strategy instances, one from each corpus, that illustrate the source-dependent
definition peak from \S\ref{sec:results-definitions}: on a noisy TCFD-CR
narrative the model is correct at D1 but routed away by keyword cues at
D2/D3, while on a TCFD-EX instance whose wording closely tracks a Recommended
Disclosure the model fails at D1 and recovers from D2 onwards.

\begin{figure*}[!tb]
\centering\small
\begin{tabular}{@{}p{0.47\textwidth}@{\hspace{0.04\textwidth}}p{0.47\textwidth}@{}}
\toprule
\textbf{Case A\,\textperiodcentered\,TCFD-CR Strategy --- D1 wins} &
\textbf{Case B\,\textperiodcentered\,TCFD-EX Strategy --- D2 onwards wins} \\
{\scriptsize rec\_test idx 119\,\textperiodcentered\,401 chars\,\textperiodcentered\,Qwen3-32B zero-shot} &
{\scriptsize disc idx 249\,\textperiodcentered\,134 chars\,\textperiodcentered\,Qwen3-32B zero-shot} \\
\addlinespace[4pt]
\textit{``In addition, the Queensland Government has established an
\$8.4 million CarbonPlus Fund administered by the department. The fund
will enable greater participation by Aboriginal landholders in carbon
markets, offset carbon emissions from the Queensland Government's car
fleet for two years, and incentivise carbon farming projects in
Queensland by valuing environmental, social and cultural
co-benefits.''}
&
\textit{``Since 2016 we have been testing the resilience of our
portfolio against the scenarios from the IEAs World Energy Outlook
(WEO) report.''}
\\
\addlinespace[6pt]
\begin{tabular}{@{}l@{\,}l@{}}
\textbf{D1} (4 pillars) & \textbf{Strategy}\,\checkmark \\
\textbf{D2} (11 RDs)    & Metrics \& Targets\,$\times$ \\
\textbf{D3} (27 criteria) & Metrics \& Targets\,$\times$ \\
\end{tabular}
&
\begin{tabular}{@{}l@{\,}l@{}}
\textbf{D1} (4 pillars) & Risk Management\,$\times$ \\
\textbf{D2} (11 RDs)    & \textbf{Strategy}\,\checkmark \\
\textbf{D3} (27 criteria) & \textbf{Strategy}\,\checkmark \\
\end{tabular}
\\
\addlinespace[6pt]
{\footnotesize Discursive narrative about a strategic
carbon-investment fund. D1 recognises the strategic intent; D2 and D3
are pulled by the dense ``carbon'' vocabulary and route to Metrics \&
Targets via keyword matching.}
&
{\footnotesize Near-verbatim paraphrase of TCFD's Strategy Recommended
Disclosure-c. D1's abstract pillar text routes to \emph{Risk
Management} on the word ``resilience''; D2 contains the matching RD
wording and maps the example directly, with D3 retaining the
criterion.}
\\
\bottomrule
\end{tabular}
\caption{Two Qwen3-32B zero-shot predictions illustrating the definition
peak-shift mechanism. Case A: a noisy TCFD-CR instance is correct at D1 but
mis-routed at D2/D3 by keyword cues. Case B: a TCFD-EX instance fails at
D1 and succeeds from D2 onwards because the matching Recommended Disclosure
wording is present at that level.}
\label{fig:def-cases}
\end{figure*}

\section{CR-specific definition control}
\label{sec:appendix-crsem20}

To test whether definitions benefit from matching the target source, we
constructed CRSEM20 from 80 labelled TCFD-CR training texts (20 per class).
It contains 270 words, the same as D2; no TCFD-CR test or TCFD-EX texts were
used. We evaluated D1, D2, and CRSEM20 with zero-shot prompting on
Llama-3.1-8B, Qwen3-8B, Qwen3.5-9B, and Qwen3-32B.

\begin{center}
\small
\setlength{\tabcolsep}{10pt}
\renewcommand{\arraystretch}{1.08}
\begin{tabular}{@{}lcc@{}}
\toprule
\textbf{Definition} & \textbf{CR} & \textbf{EX} \\
\midrule
D1 & 0.577 & 0.636 \\
D2 & 0.563 & \textbf{0.663} \\
CRSEM20 & \textbf{0.638} & 0.638 \\
\bottomrule
\end{tabular}
\captionof{table}{Mean zero-shot macro-F1 over the fixed four-model subset. Bold marks
the best definition for each evaluation source.}
\label{tab:crsem20}
\end{center}

Compared with D2, CRSEM20 improves CR by 7.5~pp but not EX ($-2.4$~pp),
suggesting that matching the target source matters beyond definition length.
Because source also changes style, content, and annotation policy, this control
does not identify which factor matters.

\section{Similarity-retrieval $k\times n$ sweep}
\label{sec:appendix-knsweep}

Table~\ref{tab:examples-knsweep} reports the full $k\times n$ sweep for
similarity retrieval.

\section{Per-base LoRA macro-F1}
\label{sec:appendix-lora-perfam}

Table~\ref{tab:appendix-lora-perfam} reports the per-base numbers behind
Figure~\ref{fig:lora-transfer-diagnosis}.

\section{Best macro-F1 per strategy}
\label{sec:appendix-best-of}

Table~\ref{tab:appendix-best-of} reports the highest macro-F1 each adaptation
strategy reaches on TCFD-CR and TCFD-EX, with the champion model in
parentheses. ClimateBERT-TCFD is included as a fully supervised reference
baseline (ClimateBERT fine-tuned on TCFD-CR training data; not part of the
eleven-model prompted portfolio).

\section{Why similarity retrieval weakens across sources}
\label{sec:appendix-retrieval-mechanism}

We first asked whether cross-source retrieval is more likely to omit the
correct class. Across six $(k,n)$ settings, this rate was similar on CR and EX
for dense retrieval (10.4\% vs. 11.2\%) and TF-IDF (10.1\% vs. 8.9\%); at
$k{=}8,n{=}2$, every prompt contained two examples per label. We then fixed
this coverage and varied the similarity of correct-class and other-class
examples separately.

Across sources, the benefit of similar correct-class examples decreases by
7.3~pp, whereas the harm from similar other-class examples changes by only
1.0~pp. The net effect of full similarity retrieval therefore falls from +7.6
to $-0.7$~pp (Table~\ref{tab:retrieval-mechanism}). The intervention uses gold
labels to define the conditions, so it is diagnostic rather than deployable.
It cannot separate style, content, and annotation-policy differences between
sources.

\begin{table*}[!tb]
\centering

\begin{minipage}[t]{0.49\textwidth}
\vspace{0pt}
\centering\small
\setlength{\tabcolsep}{5pt}
\renewcommand{\arraystretch}{1.05}
\begin{tabular}{@{}l cc cc@{}}
\toprule
& \multicolumn{2}{c}{\textbf{macro-F1}}
& \multicolumn{2}{c}{\textbf{$\Delta$ vs zero-shot (pp)}} \\
\cmidrule(lr){2-3} \cmidrule(lr){4-5}
\textbf{Configuration} & CR & EX & CR & EX \\
\midrule
Zero-shot & 0.585 & 0.643 & --- & --- \\
Random $k8,n2$ & 0.614 & 0.666 & +2.9 & +2.3 \\
\midrule
TF-IDF $k3,n2$ & 0.645 & 0.643 & +6.0 & +0.0 \\
TF-IDF $k3,n3$ & 0.654 & 0.642 & +6.9 & -0.1 \\
TF-IDF $k5,n2$ & 0.651 & 0.650 & +6.5 & +0.7 \\
TF-IDF $k5,n3$ & 0.650 & 0.642 & +6.5 & -0.1 \\
TF-IDF $k8,n2$ & 0.639 & 0.654 & +5.3 & +1.1 \\
TF-IDF $k8,n3$ & 0.651 & 0.648 & +6.6 & +0.5 \\
\midrule
Dense $k3,n2$ & 0.646 & 0.648 & +6.0 & +0.5 \\
Dense $k3,n3$ & 0.645 & 0.651 & +6.0 & +0.8 \\
Dense $k5,n2$ & 0.638 & 0.650 & +5.3 & +0.7 \\
Dense $k5,n3$ & 0.648 & 0.647 & +6.3 & +0.4 \\
Dense $k8,n2$ & 0.642 & 0.647 & +5.7 & +0.4 \\
Dense $k8,n3$ & 0.652 & 0.648 & +6.7 & +0.5 \\
\bottomrule
\end{tabular}
\captionof{table}{$k\times n$ sweep for similarity retrieval: mean macro-F1 over the
eleven prompted models for each fixed configuration, $\Delta$ vs.\ zero-shot.
$k$ caps the total retrieved examples, $n$ the per-class count; unlike
Table~\ref{tab:examples-landscape}, configurations are not tuned per model.}
\label{tab:examples-knsweep}
\end{minipage}
\hfill
\begin{minipage}[t]{0.47\textwidth}
\vspace{0pt}
\centering\small
\setlength{\tabcolsep}{3.5pt}
\renewcommand{\arraystretch}{1.08}
\begin{tabular}{@{}l cc cc cc@{}}
\toprule
 & \multicolumn{2}{c}{\textbf{Base D1}} & \multicolumn{2}{c}{\textbf{LoRA P0}} & \multicolumn{2}{c}{\textbf{LoRA D1}} \\
\cmidrule(lr){2-3} \cmidrule(lr){4-5} \cmidrule(lr){6-7}
\textbf{Base} & CR & EX & CR & EX & CR & EX \\
\midrule
Llama-3.1-8B & 0.525 & 0.605 & \textbf{0.733} & 0.666 & 0.722 & \textbf{0.670} \\
Qwen3-8B & 0.500 & 0.616 & \textbf{0.710} & 0.655 & 0.705 & \textbf{0.662} \\
Qwen3.5-9B & 0.642 & 0.661 & 0.703 & 0.652 & \textbf{0.717} & \textbf{0.673} \\
Qwen3-32B & 0.640 & 0.661 & 0.707 & 0.663 & \textbf{0.730} & \textbf{0.687} \\
\midrule
\textit{Mean} & 0.577 & 0.636 & 0.713 & 0.659 & \textbf{0.718} & \textbf{0.673} \\
\bottomrule
\end{tabular}
\captionof{table}{Per-base LoRA macro-F1, 3-seed means. \textbf{Base D1}: zero-shot
prompted, no LoRA. \textbf{LoRA P0}: adapter evaluated with the bare training
prompt. \textbf{LoRA D1}: same adapter, D1 prompt swapped in at inference time.
\textbf{Bold} marks each row's best state per setting.}
\label{tab:appendix-lora-perfam}
\end{minipage}

\vspace{1.2em}

\begin{minipage}{0.98\textwidth}
\centering\small
\setlength{\tabcolsep}{4pt}
\renewcommand{\arraystretch}{1.15}
\begin{tabular}{@{}p{0.30\textwidth} p{0.32\textwidth} p{0.32\textwidth}@{}}
\toprule
\textbf{Method} & \textbf{TCFD-CR} & \textbf{TCFD-EX} \\
\midrule
ClimateBERT-TCFD {\scriptsize (supervised)} & 0.710 & 0.648 \\
\midrule
Zero-shot + definitions (best D) & 0.657 {\scriptsize (GPT-5-mini, D1)} & 0.707 {\scriptsize (GPT-4o, D3)} \\
Random few-shot ($k{=}8,n{=}2$) & 0.689 {\scriptsize (Qwen3.5-27B)} & 0.705 {\scriptsize (Qwen3.5-27B)} \\
TF-IDF retrieval (best-on-CR) & 0.718 {\scriptsize (Llama-3.3-70B)} & \textbf{0.713} {\scriptsize (Qwen3.5-27B)} \\
Dense retrieval (best-on-CR) & 0.714 {\scriptsize (Qwen3.5-9B)} & 0.689 {\scriptsize (Qwen3.5-27B)} \\
\midrule
LoRA + P0 (training prompt) & \textbf{0.733} {\scriptsize (Llama-3.1-8B)} & 0.666 {\scriptsize (Llama-3.1-8B)} \\
LoRA + D1 (D1 swapped in) & 0.730 {\scriptsize (Qwen3-32B)} & 0.687 {\scriptsize (Qwen3-32B)} \\
\bottomrule
\end{tabular}
\captionof{table}{Best macro-F1 per strategy on TCFD-CR (in-source) and TCFD-EX
(cross-source); champion model in parentheses. \textbf{Bold} marks the
overall champion per setting.}
\label{tab:appendix-best-of}
\end{minipage}

\vspace{1.0em}

\begin{minipage}{0.98\textwidth}
\centering\small
\setlength{\tabcolsep}{8pt}
\renewcommand{\arraystretch}{1.08}
\begin{tabular}{@{}lccc@{}}
\toprule
\textbf{Controlled effect} & \textbf{In-source} & \textbf{Cross-source} &
\textbf{Cross $-$ in (95\% CI)} \\
\midrule
Similar correct-class examples & +16.7 & +9.5 & $-7.3$ [$-12.8,-1.9$] \\
Similar other-class examples & $-9.2$ & $-10.2$ & $-1.0$ [$-5.6,+3.8$] \\
Full similarity retrieval & +7.6 & $-0.7$ & $-8.3$ [$-15.4,-1.3$] \\
\bottomrule
\end{tabular}
\captionof{table}{Accuracy-point effects from the balanced intervention,
averaged over two models (Qwen3-8B and Llama-3.1-8B) and two retrievers
(TF-IDF and dense) on 112 targets (14 per source and class). Each prompt
contains eight examples, two per class.}
\label{tab:retrieval-mechanism}
\end{minipage}
\end{table*}

\end{document}